\documentclass[11pt]{article}

\usepackage[preprint]{acl}

\usepackage{times}
\usepackage{latexsym}
\usepackage[T1]{fontenc}
\usepackage[utf8]{inputenc}
\usepackage{microtype}
\usepackage{inconsolata}

\usepackage{graphicx}
\graphicspath{{figs/}}

\usepackage{tikz}
\usetikzlibrary{positioning,arrows.meta,calc,fit,backgrounds,shapes.geometric,decorations.pathreplacing}

\usepackage{amsmath}
\usepackage[ruled,vlined]{algorithm2e}
\SetKw{Continue}{continue}
\SetKw{Break}{break}
\usepackage{amsfonts}
\usepackage{amsfonts}
\usepackage{xspace}
\usepackage{amssymb}

\usepackage{nicefrac}

\usepackage{booktabs}
\usepackage{multirow}

\usepackage{enumitem}

\usepackage[table]{xcolor}
\usepackage[normalem]{ulem}

\usepackage{listings}
\lstdefinelanguage{JavaScript}{
  keywords={async, await, function, const, let, var, if, else, return, new, true, false, null, undefined},
  sensitive=true,
  comment=[l]{//},
  morecomment=[s]{/*}{*/},
  morestring=[b]",
  morestring=[b]',
}
\newcommand{\skilldisco}{\textsc{Skill-DisCo}\xspace}

\newtheorem{df}{Definition}
\newtheorem{prob}{Problem}
\newtheorem{ex}{Example}
\newtheorem{pr}{Proposition}
\newtheorem{lemm}{Lemma}
\newtheorem{lem}{Theorem}

\title{\skilldisco: Distilling and Compiling Agent Traces into Reusable Procedural Skills}

\author{
  Zhongxin Guo$^{1}$ \quad Danrui Qi$^{1}$ \quad Hanwen Gu$^{2}$ \quad Peng Cheng$^{1}$ \quad Yongqiang Xiong$^{1}$ \\
  $^{1}$Microsoft Research \quad $^{2}$Beijing Foreign Studies University \\
  \texttt{\{zhongxin.guo, danruiqi\}@microsoft.com}
}

\begin{document}
\maketitle

\begin{abstract}
Agents often repeatedly solve similar task instances from scratch,
leading to unnecessary reasoning cost and long execution traces. Prior
work has explored workflow reuse and executable skill induction, but it remains
unclear which task scenarios admit procedural skills and how the shared procedural structure should be represented across successful traces. We study this
problem in FSM-defined scenarios, where successful traces can be viewed as paths in an unknown transition graph, and formulate procedural skills as reusable parameterized control-flow subgraphs. Based on this view, we introduce \skilldisco, a
\underline{dis}tillation-and-\underline{co}mpilation framework that distills reusable PFSM subgraphs from successful traces and compiles them into callable,
executable, and verifiable procedural skills. Experiments on ALFWorld and
WebArena show that \skilldisco improves success rates and reduces agent turns
across benchmarks and model scales, demonstrating the benefits of representing
shared experience as reusable execution structures.
\end{abstract}

\section{Introduction}
\label{sec:intro}

\begin{figure*}[!t]
  \centering
  \noindent\hspace*{-1.5cm}\resizebox{\dimexpr\textwidth+1.5cm\relax}{!}{\input{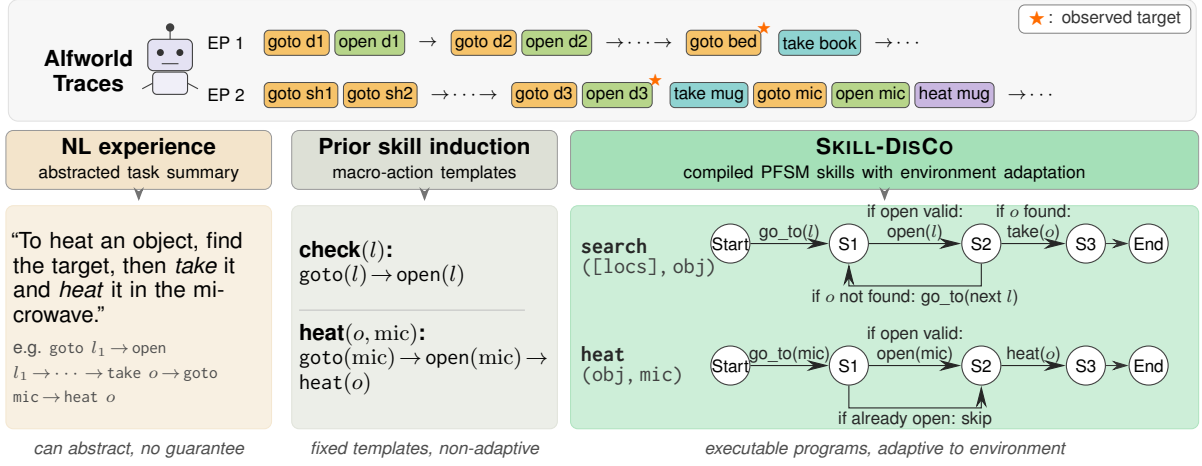}}%
  \caption{\skilldisco{} distills environment-adaptive PFSM skills that
           branch on observations, transfer across episodes and model
           scales without re-induction.}
  \label{fig:teaser}
\end{figure*}


LLM agents are increasingly used for interactive tasks that require many
reasoning and acting steps~\citep{yao2023react,shinn2023reflexion,wang2024codeact,yang2024sweagent}.
Yet even with stronger action representations such as
CodeAct~\citep{wang2024codeact}, agents often solve each task independently,
repeatedly rediscovering low-level action patterns shared across related tasks
(Figure~\ref{fig:teaser}). This increases reasoning cost, execution length, and
generalization brittleness~\citep{wang2024awm,wang2025asi}. Recent work on
experience reuse addresses this issue by extracting reusable workflows from
trajectories~\citep{wang2024awm} or inducing executable skills that improve
verifiability and composability~\citep{wang2025asi}. These results suggest that
agents should accumulate reusable procedures rather than solve each task from
scratch.

However, the foundations of procedural skill discovery remain under-specified.
Reusable skills
are meaningful when tasks share stable execution patterns, but become
ill-defined for open-ended, instance-specific generation. Existing methods
often synthesize skills directly from successful traces using LLMs. Without an
explicit notion of shared structure, the resulting libraries can become
fragmented and redundant toward trace-specific surface patterns rather
than reusable procedural logic.
We make the scope of procedural skill discovery explicit by focusing on
\emph{FSM-defined scenarios}, where execution dynamics are described by finite
states, admissible actions, and deterministic transitions. In such scenarios,
successful traces are paths in an unknown transition graph, and procedural
skills correspond to repeated transition structures that help reach goal states.
If this graph were known, task completion would reduce to graph search. In
realistic agent settings, however, the graph is unavailable, and the agent only
observes successful traces.

To capture shared structure across traces, we formalize procedural skills
through a \emph{Parameterized Finite-State Machine} (PFSM) view. A PFSM
abstracts concrete states and actions into parameterized states and operators,
so traces with different objects, states, or lengths can instantiate the same
execution pattern. Under this view, each successful trace can be lifted into a
parameterized trace graph, and a procedural skill is a reusable PFSM subgraph
matched across traces under parameter binding. Thus, a skill is not merely a
textual routine or an LLM-generated script, but a structurally grounded
abstraction of shared execution logic.


This formulation does not assume that the latent PFSM is directly observable.
In realistic agent environments, neither the full transition graph nor the
lifting function from raw traces to PFSM subgraphs is given. \skilldisco
therefore approximates PFSM-based skill discovery by recovering reusable
parameterized control-flow patterns from successful traces and validating them
through skill compilation.
This yields a distillation-and-compilation framework. Distillation discovers
compact, high-coverage PFSM subgraphs that favor reusable structures over
trace-specific routines. Compilation turns each discovered structure into a
callable, executable, and verifiable skill through explicit specification and
execution-grounded verification.

Our contributions are as follows:

\textbf{\textit{(C1)}} We make explicit the scope of procedural skill discovery by focusing on FSM-defined scenarios, where successful traces are paths in an unknown transition graph and reusable skills have well-defined transition semantics.

\textbf{\textit{(C2)}} We formulate procedural skills as reusable parameterized
control-flow subgraphs that can match multiple successful traces under
parameter binding. This provides structural targets for skill discovery,
rather than treating each successful trace as an independent routine.

\textbf{\textit{(C3)}} We introduce \skilldisco, a distillation-and-compilation framework that approximates these structural targets and compiles them into verified and executable procedural skills.

\textbf{\textit{(C4)}} Extensive experiments show that \skilldisco improves both
success rate and efficiency across different benchmarks and model scales. Further
analysis shows that compact compiled skills reduce library redundancy,
improve execution reliability, and transfer procedural knowledge from
stronger induction models to smaller execution models.

\section{Preliminaries}
\label{section: problem}
\setlength{\abovedisplayskip}{3pt plus 1pt minus 1pt}
\setlength{\belowdisplayskip}{3pt plus 1pt minus 1pt}
\setlength{\abovedisplayshortskip}{2pt plus 1pt minus 1pt}
\setlength{\belowdisplayshortskip}{2pt plus 1pt minus 1pt}

\subsection{FSM-Defined Scenarios}
 In this paper, we focus on scenarios whose execution dynamics can be formulated as finite-state machines. Such scenarios have a finite state space, a finite action space, and deterministic transitions: given a state $s \in \mathcal{S}$ and an action $a \in \mathcal{A}$, the next state is uniquely determined by $\delta(s,a)$. We refer to such scenarios as \textit{FSM-defined scenarios}.

\begin{df}[FSM-Defined Scenario]

A scenario is called an \textit{FSM-defined scenario} if its execution dynamics
can be represented as a finite-state machine
\[
\mathcal{M} = (\mathcal{S}, \mathcal{A}, \delta, \mathcal{S}_0, \mathcal{S}_{\mathrm{goal}}),
\]
where $\mathcal{S}$ is a finite set of states, $\mathcal{A}$ is a finite set of
actions, $\delta: \mathcal{S} \times \mathcal{A} \rightarrow \mathcal{S}$ is a
deterministic transition function, $\mathcal{S}_0 \subseteq \mathcal{S}$ is the
set of possible initial states, and $\mathcal{S}_{\mathrm{goal}} \subseteq
\mathcal{S}$ is the set of goal states.
The corresponding transition graph is defined as
$
G^{*} = (V^{*}, E^{*})$,
where $V^{*} = \mathcal{S}$ and
$E^{*}
=
\{(s,a,s') \mid s,s' \in \mathcal{S}, a \in \mathcal{A},
\delta(s,a)=s'\}.$
\end{df}

For an FSM-defined scenario, if the complete transition graph $G^{*}$ and the goal states are known, task completion can be reduced to finding a path from an initial state to a goal state on $G^{*}$. However, the complete transition graph is unavailable in many agent settings. Thus, we aim to infer reusable transition structures from past successful traces.
\begin{df}[Primitive Operator]
A primitive operator is a tuple
\[
op=(\mathcal{X}, \mathcal{Y}, \mathrm{Pre}, \mathrm{Post}),
\],
where $\mathcal{X}$ is the input space, $\mathcal{Y}$ is the output space,
$\mathrm{Pre}: \mathcal{X}\rightarrow\{0,1\}$ is a pre-condition predicate, and
$\mathrm{Post}: \mathcal{X}\times\mathcal{Y}\rightarrow\{0,1\}$ is a
post-evaluation predicate.
For any input $x\in\mathcal{X}$ satisfying $\mathrm{Pre}(x)=1$, executing $op$
produces a unique output $y=op(x)\in\mathcal{Y}$ such that $
\mathrm{Post}(x,y)=1.
$
\end{df}

\begin{df}[Successful Trace]
Given an FSM-defined scenario
$\mathcal{M} = (\mathcal{S}, \mathcal{A}, \delta, \mathcal{S}_0, \mathcal{S}_{\mathrm{goal}})$, an agent trace is a finite execution record
$\tau =
(o_0, a_0, o_1, a_1, \ldots, a_{T-1}, o_T)$,
where $o_t$ denotes the observation received by the agent at step $t$ and
$a_t \in \mathcal{A}$ is the action executed by the agent.

The trace is induced by an underlying state trajectory
$
(s_0, a_0, s_1, a_1, \ldots, a_{T-1}, s_T),
$
such that
$
s_{t+1} = \delta(s_t, a_t),
$
and each observation $o_t$ is generated from the underlying state $s_t$.
We call $\tau$ a successful agent trace if the underlying execution starts from
a valid initial state and reaches a goal state, i.e. $
s_0 \in \mathcal{S}_0$ and $
s_T \in \mathcal{S}_{\mathrm{goal}}.$
\end{df}

Note that every action $a \in \mathcal{A}$ in an FSM-defined scenario is treated as a \textit{Primitive Operator}, i.e. actions in a \textit{Successful Trace} are all primitive operators. For clarity, we will simply refer to such primitive operators as actions throughout the paper.

\subsection{PFSM-Based Procedural Skill Discovery}

A standard FSM represents task completion as a concrete path from an initial
state to a goal state. Such a path precisely records one execution, but is often
too instance-specific because each transition is grounded in concrete states and
actions. As a result, executions with the same control logic may appear as
different paths when they involve different objects, locations, or intermediate
states.

To capture reusable execution patterns, we introduce the notion of a
\textit{Parameterized Finite State Machine} (PFSM). A PFSM abstracts concrete
states and actions with parameters, allowing multiple concrete FSM paths to be
represented by the same parameterized control-flow structure.

\begin{df}[Parameterized FSM (PFSM)]

A \textit{Parameterized Finite State Machine} is defined as
\[
\widetilde{\mathcal{M}}
=
(\widetilde{\mathcal{S}}, \widetilde{\mathcal{A}}, \Theta, \widetilde{\delta},
\widetilde{\mathcal{S}}_0, \widetilde{\mathcal{S}}_{\mathrm{goal}}),
\]
where $\widetilde{\mathcal{S}}$ is a finite set of parameterized states,
$\widetilde{\mathcal{A}}$ is a finite set of parameterized actions,
$\Theta$ is the parameter space, and
$
\widetilde{\delta}:
\widetilde{\mathcal{S}} \times \widetilde{\mathcal{A}} \times \Theta
\rightarrow
\widetilde{\mathcal{S}}
$
is a deterministic parameterized transition function. Each parameter assignment
$\theta \in \Theta$ instantiates a parameterized transition into a concrete FSM
transition.
Here, a parameterized state $\tilde{s}\in\widetilde{\mathcal{S}}$ represents an
abstract execution state rather than a fully grounded environment state. A
parameterized action $\tilde{a}\in\widetilde{\mathcal{A}}$ represents an action
schema whose arguments are instantiated by parameters.
\end{df}

\paragraph{Example 1.}
Consider two successful traces for finding and taking a target object:

\vspace{2pt}
{\raggedright
\noindent\hangindent=1.8em\hangafter=1
\makebox[1.8em][l]{$\tau_1$:}%
\texttt{go\_to(drawer1)}\,$\rightarrow$\,\allowbreak
\texttt{open(drawer1)}\,$\rightarrow$\,\allowbreak
\texttt{go\_to(drawer2)}\,$\rightarrow$\,\allowbreak
\texttt{open(drawer2)}\,$\rightarrow$\,\allowbreak
\texttt{go\_to(desk1)}\,$\rightarrow$\,\allowbreak
\texttt{go\_to(bed)}\,$\rightarrow$\,\allowbreak
\texttt{take(book)}\,$\rightarrow$\,\allowbreak
\texttt{end}.\par}

\vspace{1pt}
{\raggedright
\noindent\hangindent=1.8em\hangafter=1
\makebox[1.8em][l]{$\tau_2$:}%
\texttt{go\_to(shelf1)}\,$\rightarrow$\,\allowbreak
\texttt{go\_to(shelf2)}\,$\rightarrow$\,\allowbreak
\texttt{go\_to(drawer1)}\,$\rightarrow$\,\allowbreak
\texttt{open(drawer1)}\,$\rightarrow$\,\allowbreak
\texttt{go\_to(drawer2)}\,$\rightarrow$\,\allowbreak
\texttt{open(drawer2)}\,$\rightarrow$\,\allowbreak
\texttt{take(mug)}\,$\rightarrow$\,\allowbreak
\texttt{end}.\par}

\vspace{2pt}
\noindent
In a standard FSM, these two traces correspond to different concrete paths,
because they visit different numbers of locations. However, they share the same
parameterized control-flow pattern:

\vspace{2pt}
\noindent
\hspace*{1.5em}\begin{tabular}{@{}l@{\hskip 1.5em}l@{}}
\textbf{repeat} & \texttt{go\_to}($l$) \\
                & \textbf{if} \texttt{can\_open}($l$) \textbf{then} \texttt{open}($l$) \\
                & \texttt{check\_target}($l, o$) \\
\textbf{until} & \texttt{found}($o$) \\
\textbf{then}  & \texttt{take}($o$).
\end{tabular}

This pattern can be represented as a PFSM with parameterized actions such as
$\texttt{go\_to}(l)$ and $\texttt{take}(o)$, where $l$ is a location parameter
and $o$ is an object parameter. The loop over candidate locations abstracts away
the number and identity of concrete locations, while preserving the reusable
control logic needed to complete the task.

Given a set of successful traces, each trace can be lifted from a concrete
FSM path to a subgraph of the PFSM transition graph. Such a subgraph captures
the parameterized execution structure of a successful trace, including its abstract states,
parameterized actions, and control-flow relations. We refer to this lifted
subgraph as a \textit{parameterized trace graph}. Since the underlying PFSM is unavailable, the lifting function $\phi: \tau_i \mapsto \widetilde{G}_i$ does not admit a closed-form definition. We approximate it via a multi-stage pipeline described in Section~\ref{sec:method}.

Our goal is to discover procedural skills from successful traces.
Intuitively, a procedural skill corresponds to a PFSM subgraph that is shared
by many successful traces. Such a subgraph captures a recurring control-flow
structure that helps the agent reach a goal across different concrete
instantiations.

\begin{prob}[Procedural Skill Discovery]

Given a set of successful traces $\mathcal{T}^{+}=\{\tau_1,\tau_2,\ldots,\tau_N\}$ and a lifting function $\phi:\tau_i \mapsto \widetilde{G}_i$ that maps each trace to a parameterized trace graph, the goal is to discover a set of procedural skills $\mathcal{K}=\{K_1,K_2,\ldots,K_m\}$, where each skill $K_j$ is a parameterized control-flow subgraph that matches a subset of $\{\widetilde{G}_i\}_{i=1}^N$ under parameter binding, i.e. $K_j \preceq \widetilde{G}_i$ for $i$ in some index set.
\end{prob}

Intuitively, a desirable skill set $\mathcal{K}$ should satisfy three
properties: \textit{\textbf{(i) Coverage}}, where each skill is supported by
multiple successful traces rather than a single execution, \textit{\textbf{(ii)
Utility}}, where skills capture control-flow structures that help reach goal
states, and \textit{\textbf{(iii) Compactness}}, where skills are neither overly
specific nor trivially generic. In realistic agent settings, neither the full
PFSM $\widetilde{\mathcal{M}}$ nor the lifting function $\phi$ is analytically available.
\skilldisco approximates both from successful traces through the pipeline in
Section~\ref{sec:method}.


\section{The \skilldisco Framework}
\label{sec:method}

Section~\ref{section: problem} formulates procedural skill discovery as
identifying reusable PFSM subgraphs from successful agent traces. Solving this
problem directly is challenging because the complete PFSM
$\widetilde{\mathcal{M}}$ is unavailable, the lifting function $\phi$ from raw
traces to parameterized trace graphs must be inferred, and exact subgraph
matching under parameter binding is costly and noise-sensitive. To address
these challenges, \skilldisco discovers procedural skills through distillation
and compilation.

\subsection{Framework Overview}
\label{sec:skilldisco-overview}

\begin{figure*}[t]
    \centering
      \noindent\hspace*{-1.5cm}\resizebox{\dimexpr\textwidth+1.5cm\relax}{!}{\input{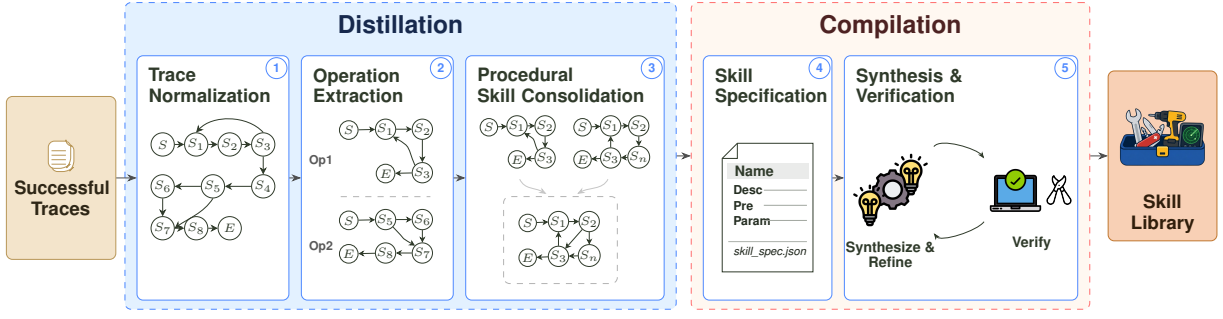}}
    \caption{Overview of \skilldisco. \emph{Distillation} phase turns successful traces into reusable PFSM subgraphs; \emph{Compilation} phase converts them into executable and verified skills.}
    \label{fig:overview}
\end{figure*}

Figure~\ref{fig:overview} shows an overview of the \skilldisco framework. The input
to \skilldisco is a set of successful traces $\mathcal{T}^{+}=\{\tau_1,\tau_2,\ldots,\tau_N\}$, where $N$ denotes the number of successful
traces. The output is a skill library
$\mathcal{K}=\{K_1,K_2,\ldots,K_m\}$, which contains the $m$ procedural skills.

\skilldisco contains two phases. The distillation phase approximates the
latent lifting function from traces to reusable PFSM subgraphs. It first
normalizes each raw trace into an executable intermediate program, then
segments the program into subgoal-level operations, and finally
clusters fragments that share parameterized control-flow structure. Each
cluster is treated as an approximate PFSM subgraph and becomes a skill
candidate.

The compilation phase turns each candidate structure into an executable
artifact. It first constructs a skill specification with a signature,
description, behavioral requirements, and metadata. It then synthesizes Python
code and verifies it on held-out tasks; only skills that pass verification are
included in the final callable library.

\subsection{The Distillation Phase}
\label{sec:distillation-phase}

The distillation phase consists of three stages that jointly approximate the latent lifting function
$\phi: \tau_i \mapsto \widetilde{G}_i$ and discover reusable PFSM subgraphs from successful
traces. As described in Section~\ref{section: problem}, each raw execution trace is first
abstracted into a concrete trace graph over the FSM. This concrete trace graph is then lifted
into a PFSM graph, where repeated action patterns are represented as parameterized,
code-like control-flow structures with branches. Finally, reusable subgraphs are identified
from the resulting PFSM graphs and extracted as procedural skills.
Therefore, the lifting function $\phi$ is not implemented as a single monolithic step. Instead,
it is decomposed into three successive transformations, each corresponding to one stage of
the distillation phase.

\textbf{Stage 1: Trace Normalization.}
Trace normalization converts each successful raw trace into a structured,
executable intermediate representation. Raw agent logs interleave reasoning
text, tool calls, observations, and action histories, but do not expose the
control-flow structure needed for PFSM-based skill discovery. \skilldisco
therefore normalizes each trace into a program $p_i$ that preserves primitive
operators and observations while making loops, branches, and parameterized
entities explicit.

Each normalized program has three components: primitive environment calls paired
with observations, symbolic variables for task-specific entities such as
objects or web elements, and code-level control flow for repeated action
patterns and observation-conditioned decisions. For example, a trace that visits
candidate locations until an object is found can become a loop whose guard
depends on the current observation.

In our implementation, an LLM extracts all executed actions and observations,
preserves their order, replaces concrete entities with typed variables when
possible, and emits a Python-like program. This program is not the final skill;
it is an intermediate representation from which reusable subgoal-level
operations are extracted.


\textbf{Stage 2: Subgoal-level Operation Extraction.}
The goal of Stage~2 is to decompose each normalized program into
subgoal-level operations that can serve as candidate procedural skills. A
complete trace often contains multiple reusable parts. For example, a
single household task may include searching for an object, retrieving it,
navigating to an appliance, and applying the appliance. Treating the whole
trace as one skill would make the induced routine too specific, while
treating each primitive operator as a skill would lose the benefit of
procedural reuse. \skilldisco therefore extracts intermediate-granularity operators that correspond to coherent subgoals.

Given a normalized program $p_i$, the operator extractor emits
$
\mathcal{O}_i=\{o^{(i)}_1,\ldots,o^{(i)}_{m_i}\},
$
where each operator is represented as
$
o=(\nu,\sigma,\mathbf{u},c).
$
Here, $\nu$ is an action-oriented identifier, $\sigma$ is a natural-language
summary, $\mathbf{u}=(u_1,\ldots,u_{|o|})$ is the primitive operator sequence, and $c$ is the corresponding code fragment. We discard unit-length
fragments with $|o|=1$, since they correspond to primitive operators rather
than reusable procedural structure. The remaining multi-step operators are
collected as
$
\mathcal{O}_{\mathrm{multi}}
=
\bigcup_i \{o\in\mathcal{O}_i: |o|\geq 2\}.
$

\textbf{Stage 3: Procedural Skill Consolidation.}
Stage~3 consolidates procedural skills by clustering subgoal-level operations that
approximate the same reusable PFSM subgraph. Intuitively, a useful procedural
skill should be supported by multiple successful traces, rather than being an
incidental pattern that appears in only one execution.

In \skilldisco, subgoal-level operations are grouped together when they share
the same underlying parameterized execution structure, even if their concrete
objects, locations, or action lengths differ. For example, the two traces in Example~1 differ in their concrete locations,
objects, and path lengths, but share the same parameterized search-and-take
control flow. 

Each cluster approximates a reusable PFSM subgraph and receives a reusability
score:
\[
r_k
\approx
\frac{1}{N}
\sum_{i=1}^{N}
\mathbb{I}[K_k \preceq \widetilde{G}_i],
\]
where $K_k \preceq \widetilde{G}_i$ indicates that the PFSM subgraph
approximated by cluster $c_k$ can be matched to the lifted trace graph
$\widetilde{G}_i$ under some parameter binding. The score is estimated from
trace coverage and operation statistics. For each high-coverage cluster, \skilldisco abstracts the shared
subgoal-level operation as a procedural skill.

\subsection{The Compilation Phase}
\label{sec:compilation}

The distillation phase abstracts high-coverage clusters into procedural skills representing reusable PFSM subgraphs. However, a PFSM subgraph is only a
structural artifact, i.e. it does not define a callable interface, runtime grounding rules, return protocol, or failure handling. The compilation phase turns each procedural skill into a callable, executable, and verifiable version.
In \skilldisco, the compilation phase consists of two stages: \textit{skill specification} and \textit{skill synthesis and verification}.

\textbf{Stage 4: Skill Specification.}
Stage~4 augments discovered procedural skills with explicit specifications.
For each skill abstracted from a high-coverage cluster, \skilldisco
creates a skill definition that captures how the skill should be invoked
and what behavior it should satisfy before any concrete implementation is
generated. Each skill definition contains:
(i)~a \emph{signature}, including an action-oriented skill name, typed
parameters with default values, and a structured return type 
(ii)~a \emph{description}, serving as both a machine-readable docstring and
LLM-facing guidance 
(iii)~\emph{behavioral requirements}, including preconditions, postconditions,
and declared side effects and
(iv)~\emph{metadata}, including the skill abstraction level, expected primitive
actions amortized per invocation, and a confidence score inherited from $r_k$.

The explicit specification guides implementation synthesis and provides
concrete criteria for verification, while giving the deployed agent a
domain-agnostic vocabulary for selecting skills during task execution.

\textbf{Stage 5: Synthesis and Verification.}
Stage~5 turns each skill specification into a verified executable skill through
a synthesis and verification loop. Given a skill definition, \skilldisco
first synthesizes a Python program that realizes the
specified behavior using the primitive actions. During execution, $f_k$ interacts with the workload environment and returns a
structured output that provides execution information to the caller.

On a held-out set, verification checks runtime correctness, postcondition
satisfaction, and action savings. Skills that fail are re-synthesized with
feedback for up to $R$ retries; any remaining failures are discarded. 

The result is a verified executable skill library: each skill exposes a
documented signature that agents can inspect, select, and call during task
execution.

\section{Experiments}
\label{sec:experiments}

We evaluate whether \skilldisco can distill reusable procedural skills from
successful traces and deploy them as reliable skills for interactive agents.
Our experiments are organized around four questions: \textit{\textbf{RQ1.}} Does \skilldisco improve end-to-end task success rate and turn efficiency compared with base agents and prior skill-induction methods?
(\S\ref{sec:rq1}) \textit{\textbf{RQ2.}} Do the procedural skills transfer across model families, scales, and
reasoning modes? (\S\ref{sec:rq2}) \textit{\textbf{RQ3.}} How are the learned skills invoked and executed at inference time? (\S\ref{sec:rq3}) \textit{\textbf{RQ4.}} Do all components of \skilldisco contribute to  the observed gains? (\S\ref{sec:ablations})

\subsection{Experimental Setup}
\label{sec:exp-setup}


\textbf{Datasets.}
We evaluate on two interactive agent datasets with different execution
structures: \textit{(1) ALFWorld}~\citep{shridhar2020alfworld} consists of
text-based household tasks that require navigation, search, and object
manipulation. \textit{(2) WebArena}~\citep{zhou2024webarena} consists of
realistic web-navigation tasks over self-hosted websites.

\textbf{Splits.}
For each dataset, we use a strict \emph{induction/evaluation} split: skills are
derived only from the induction split and evaluated only on held-out tasks.
For \textit{ALFWorld}, the induction split consists of 200 tasks sampled from the
official \texttt{train} set, and the evaluation split is the official unseen
split with 134 tasks. For WebArena, which has no
canonical induction split, we split the 812 tasks in half: the first 406 tasks
for induction and the remaining 406 for evaluation.
All baselines use the same splits as \skilldisco.

\textbf{Baselines.}
We compare against two categories of baselines. The first category contains
base agents without skill augmentation: ReAct~\citep{yao2023react} using the
AgentBench implementation~\citep{liu2023agentbench}, and
CodeAct~\citep{wang2024codeact}. The second group consists of offline workflow-
and skill-induction methods: AWM\textsubscript{offline}~\citep{wang2024awm} and
the offline variant of ASI, ASI\textsubscript{offline}~\citep{wang2025asi}.
For all skill-augmented variants, each skill's signature and description are
appended in agents' prompts to instruct LLMs to decide whether and how to
invoke a skill or emit a primitive action.


\textbf{Metrics.}
We report \emph{success rate} (SR) and \emph{average
agent turns} (Avg. Turns) per episode. 
Full per-episode token usage and inference cost are reported in
Appendix~\ref{sec:appendix-token-detail}.


\subsection{End-to-End Performance}
\label{sec:rq1}

We first evaluate whether \skilldisco improves complete task execution when
added to existing interactive agents. Table~\ref{tab:baseline-comparison}
compares \skilldisco with base agents and prior offline skill-induction
methods using GPT-4o. \skilldisco also uses
GPT-4o to induce the skill library.

\begin{table}[t]
\centering
\small
\setlength{\tabcolsep}{3pt}
\caption{End-to-end results on ALFWorld and WebArena.
Parentheses show relative changes over the base agent; \textbf{bold} marks the best result and \underline{underlining} marks the second-best.}
\label{tab:baseline-comparison}

\begin{tabular}{@{}l c c@{}}
\toprule
\textbf{Method} & \textbf{SR (\%)} $\uparrow$ & \textbf{Avg.\ Turns} $\downarrow$ \\
\midrule
\multicolumn{3}{@{}l}{\textit{(a) ALFWorld}} \\
ReAct                         & 82.0          & 19.3 \\
CodeAct                       & \underline{96.3} & \underline{3.6} \\
\cmidrule(l){2-3}
AWM\textsubscript{offline}    & 54.5          & 11.3 \\
ASI\textsubscript{offline}    & 47.0          & 11.4 \\
\cmidrule(l){2-3}
\textbf{\skilldisco\,$+$\,ReAct}
  & 92.4\,{\tiny($+12.7\%$)} & 8.8\,{\tiny($-54.5\%$)} \\
\textbf{\skilldisco\,$+$\,CodeAct}
  & \textbf{99.3}\,{\tiny($+3.1\%$)} & \textbf{3.2}\,{\tiny($-11.3\%$)} \\
\midrule
\multicolumn{3}{@{}l}{\textit{(b) WebArena}} \\
ReAct                         & 23.9          & 5.9 \\
CodeAct                       & 20.0          & 10.9 \\
\cmidrule(l){2-3}
AWM\textsubscript{offline}    & 21.2          & 5.9 \\
ASI\textsubscript{offline}    & \underline{24.6} & \underline{5.7} \\
\cmidrule(l){2-3}
\textbf{\skilldisco\,$+$\,ReAct}
  & \textbf{29.1}\,{\tiny($+21.6\%$)} & \textbf{5.1}\,{\tiny($-13.1\%$)} \\
\textbf{\skilldisco\,$+$\,CodeAct}
  & 22.9\,{\tiny($+14.8\%$)} & 8.5\,{\tiny($-22.0\%$)} \\
\bottomrule
\end{tabular}
\end{table}

\textbf{Higher task success rate.}
\skilldisco achieves the highest SR on all benchmarks under our evaluation
setting. On \textit{ALFWorld}, \skilldisco\,$+$\,CodeAct improves over the strong
CodeAct from 96.3\% to 99.3\%. On WebArena,
\skilldisco\,$+$\,ReAct improves ReAct from 23.9\% to 29.1\%, outperforming
ASI\textsubscript{offline} by 4.5 absolute points. Since all offline
skill-induction methods use the same induction and evaluation splits, these
gains reflect the quality of the induced skills by \skilldisco.

\textbf{Consistent reduction in agent turns.}
\skilldisco also reduces average turns across all settings: $-54.5\%$/$-11.3\%$
(ReAct/CodeAct) on \textit{ALFWorld} and $-13.1\%$/$-22.0\%$ on WebArena. By
delegating repeated low-level action sequences to the reliable procedural skills, agents solve each task with fewer decisions.

\textbf{Complementary to different agents.}
Table~\ref{tab:baseline-comparison} shows that the same library improves both
ReAct (natural-language reasoning) and CodeAct (code-emitting) across all
benchmarks, indicating that \skilldisco acts as an agent-agnostic augmentation layer by providing reusable procedural skills, rather than being tied to a specific agent.

\begin{table}[t]
\centering
\small
\setlength{\tabcolsep}{3pt}
\caption{Skill transfer across target models on ALFWorld and WebArena.
$\Delta$ denotes relative change over the vanilla setting; \textbf{bold} marks the best result.}
\label{tab:multi-models}

\begin{tabular}{@{}l c c c c@{}}
\toprule
 & \multicolumn{2}{c}{\textbf{SR (\%)} $\uparrow$} & \multicolumn{2}{c}{\textbf{Avg.\ Turns} $\downarrow$} \\
\cmidrule(lr){2-3}\cmidrule(lr){4-5}
\textbf{Model} & \textbf{Van.} & \textbf{+Sk.\,($\Delta$)} & \textbf{Van.} & \textbf{+Sk.\,($\Delta$)} \\
\midrule
\multicolumn{5}{@{}l}{\textit{(a) ALFWorld}} \\
Qwen3.5-4B        & 61.2 & 91.0\,{\tiny($+48.8\%$)} & \phantom{0}9.0 & 5.0\,{\tiny($-44.4\%$)} \\
Qwen3.5-9B        & 54.5 & 98.5\,{\tiny($+80.8\%$)} & 10.8 & 3.4\,{\tiny($-68.2\%$)} \\
GPT-4o            & 96.3 & \textbf{99.3}\,{\tiny($+3.1\%$)}  & \phantom{0}3.6 & 3.2\,{\tiny($-11.3\%$)} \\
GPT-4o-mini       & 71.6 & 94.8\,{\tiny($+32.3\%$)} & \phantom{0}7.1 & 4.4\,{\tiny($-38.0\%$)} \\
GPT-5-chat        & 91.0 & 98.5\,{\tiny($+8.2\%$)}  & \phantom{0}4.7 & \textbf{3.0}\,{\tiny($-34.6\%$)} \\
GPT-5-mini        & 91.8 & 98.5\,{\tiny($+7.3\%$)}  & \phantom{0}4.9 & 3.3\,{\tiny($-33.3\%$)} \\
\midrule
\multicolumn{5}{@{}l}{\textit{(b) WebArena}} \\
Qwen3.5-4B        & 10.1 & 18.7\,{\tiny($+85.3\%$)} & 8.6 & 6.8\,{\tiny($-20.4\%$)} \\
Qwen3.5-9B        & 21.2 & 23.9\,{\tiny($+12.8\%$)} & 6.8 & 6.0\,{\tiny($-12.0\%$)} \\
GPT-4o            & 23.9 & 29.1\,{\tiny($+21.6\%$)} & 5.9 & 5.1\,{\tiny($-13.1\%$)} \\
GPT-4o-mini       & 18.0 & 18.0\,{\tiny($\pm0.0\%$)} & 7.7 & 7.1\,{\tiny($-7.7\%$)} \\
GPT-5-chat        & 32.5 & \textbf{37.0}\,{\tiny($+13.7\%$)} & 5.3 & \textbf{4.8}\,{\tiny($-8.6\%$)} \\
GPT-5-mini        & 32.5 & 33.7\,{\tiny($+3.8\%$)}  & 5.4 & 4.8\,{\tiny($-10.1\%$)} \\
\bottomrule
\end{tabular}
\end{table}

\begin{table*}[t]
\centering
\small
\setlength{\tabcolsep}{6pt}
\caption{Skill-usage statistics.
\#Sk is the number of induced skills; Turns and Sk.\ turns are average total and
skill-invoking turns; Avg./Max prim. steps are primitive steps collapsed per skill
call; Err. is execution-error rate.}
\label{tab:skill-usage}

\begin{tabular}{@{}l l c c c c c c@{}}
\toprule
\textbf{Bench} & \textbf{Method} & \textbf{\#Sk} & \textbf{Turns} $\downarrow$ & \textbf{Sk.\ turns} & \textbf{Avg.\ prim.\ steps} & \textbf{Max prim.\ steps} & \textbf{Err. (\%)} $\downarrow$ \\
\midrule
\multirow{2}{*}{ALFWorld}
 & ASI\textsubscript{offline}    & 110   & 11.4 & 1.4 & 0.2 & 10 & 75.3 \\
 & \textbf{\skilldisco}          & 5     & \textbf{3.2} & 2.6 & 5.6 & 33 & \textbf{0.0} \\
\midrule
\multirow{2}{*}{WebArena}
 & ASI\textsubscript{offline}    & 146   & 5.5 & 1.3 & 2.5 & \phantom{0}8 & 33.9 \\
 & \textbf{\skilldisco}          & 20    & \textbf{5.1} & 1.0 & 2.8 & 36 & \textbf{21.5} \\
\bottomrule
\end{tabular}
\end{table*}



\subsection{Cross-Model Skill Transferability }
\label{sec:rq2}

We next test whether the skill library induced by \skilldisco with GPT-4o can
transfer to other models. We deploy the same skill library \emph{without modification}
across multiple target models: Qwen3.5-4B, Qwen3.5-9B, GPT-4o, GPT-4o-mini,
GPT-5-chat, and GPT-5-mini. 
Based on Table~\ref{tab:baseline-comparison}, we use the skill library induced from the stronger \skilldisco configuration on each benchmark: CodeAct for \textit{ALFWorld} and ReAct for \textit{WebArena}.


\textbf{Skills transfer across target models.}
With the same skill library, every target model benefits:
average turns drop in every cell ($-7.7\%$ to $-68.2\%$), while SR improves by
$+3.1\%$ to $+80.8\%$ on \textit{ALFWorld} and by up to $+85.3\%$ on WebArena
with no decreases. These results show that the induced skills are robust
across model families and scales.

\textbf{Weaker models can benefit substantially from skill transfer.}
Although gains are not strictly monotone in capacity, the largest jumps
occur on the smaller open-source backbones: Qwen3.5-9B gains $+80.8\%$ on
\textit{ALFWorld} and Qwen3.5-4B gains $+85.3\%$ on WebArena, versus only
$+3.1\%$--$+7.3\%$ for GPT-4o/GPT-5-mini. Strikingly, Qwen3.5-9B with the
GPT-4o-induced library reaches 98.5\% on \textit{ALFWorld}, surpassing its
inducer (96.3\%) at a fraction of the cost. This shows that \skilldisco
serves as a form of \emph{procedural-knowledge distillation and transfer}, allowing
smaller models to benefit from skills compiled by a stronger inducer.

\subsection{Skill Usage and Execution Reliability}
\label{sec:rq3}

We analyze how much of an episode the agent delegates to its skill library and
how reliably those skill calls execute. Following the settings in
Table~\ref{tab:baseline-comparison} (CodeAct on \textit{ALFWorld} and ReAct on
WebArena), we report aggregate skill-usage statistics in
Table~\ref{tab:skill-usage} and compare \skilldisco with
ASI\textsubscript{offline}.

\textbf{\skilldisco learns compact, high-coverage skills.}
\skilldisco uses far fewer skills than ASI (5 vs.\ 110 on \textit{ALFWorld};
20 vs.\ 146 on \textit{WebArena}) while each call covers more prim. steps
(5.6 vs.\ 0.2 on \textit{ALFWorld}; 2.8 vs.\ 2.5 on \textit{WebArena}) and
larger maximum step compression (33/36 vs.\ 10/8). Thus, \skilldisco distills compact
skills that encode reusable procedures rather than shallow fragments.

\textbf{Verified skills execute more reliably.}
\skilldisco makes skill calls less error-prone, reducing execution errors from
75.3\% to 0.0\% on \textit{ALFWorld} and from 33.9\% to 21.5\% on \textit{WebArena}.
Thus, its gains come not only from invoking skills, but from invoking skills
that execute reliably; together with higher step compression, this explains
the SR and turn improvements in Table~\ref{tab:baseline-comparison}.

\subsection{Ablation Study}
\label{sec:ablations}

We evaluate the contribution of the two main phases of \skilldisco,
distillation and compilation, through ablations on ALFWorld using CodeAct with
GPT-4o (Table~\ref{tab:ablations}).

\begin{table}[t]
\centering
\caption{Pipeline ablations.
\#Sk = skill library size;
Relative SR $\Delta$ vs.\ the full pipeline in parentheses.}
\label{tab:ablations}
\small\setlength{\tabcolsep}{4pt}
\begin{tabular}{@{}l r r r@{}}
\toprule
\textbf{Configuration} & \textbf{\#Sk} & \textbf{SR (\%)} & \textbf{Turns} \\
\midrule
A1: no distill & 43 & 53.0\,{\scriptsize($-46.6\%$)} & 11.5 \\
A2: no compile & NA  & 97.0\,{\scriptsize($-2.3\%$)}  & 3.6 \\
\midrule
Full \skilldisco  & 5  & \textbf{99.3}                & \textbf{3.2} \\
\bottomrule
\end{tabular}
\end{table}

\textbf{Distillation is the main driver of success.}
Removing distillation and inducing skills per successful trace yields a
larger 43-skill library but reduces SR from 99.3\% to 53.0\% and increases
turns from 3.2 to 11.5. Without cross-trace consolidation the
library is dominated by overlapping, trace-specific variants, making skill
selection harder and causing wrong-variant invocations or fallback to long
open-loop executions.

\textbf{Compilation mainly improves execution compactness.}
Removing compilation and shipping the output of Stage~3 as natural-language
procedure descriptions only modestly reduces SR ($99.3\!\to\!97.0\%$), but each
episode must keep the full procedure text in context, inflating tokens
(Appendix~\ref{sec:appendix-token-detail}). Compiling skills into callable
code externalizes procedural execution from the prompt, yielding both
efficiency and a small additional reliability gain.



\section{Related Work}
\label{sec:related}

\textbf{LLM-Based Agents.}
ReAct~\citep{yao2023react} interleaves reasoning with actions;
CodeAct~\citep{wang2024codeact}, PAL~\citep{gao2023pal}, and Code as
Policies~\citep{liang2023code} show executable code improves compositional
reasoning~\citep{wei2022cot}. Yet procedural knowledge in successful traces
remains ephemeral and rarely reused.

\textbf{Reusing Agent Experience as Behavioral Imitation.}
A common line keeps the model fixed and reuses experience as
\emph{token-level context}: Reflexion~\citep{shinn2023reflexion} prompts
self-reflection; AWM~\citep{wang2024awm} induces textual workflows;
Trace2Skill~\citep{trace2skill2026} writes structured skill documents.
All store knowledge in \emph{natural language}---unverifiable before
deployment and re-interpreted at inference. \skilldisco instead compiles
procedural knowledge into external executable code, verifiable offline and
reusable across models.

\textbf{Synthesizing Executable Tools, Workflows, and Skills.}
\textit{(1) Tool and workflow generation.}
LATM~\citep{cai2023toolmakers}, CREATOR~\citep{qian2023creator}, and
ToolGen~\citep{wang2024toolgen} synthesize or unify tools per task;
AgentDistill~\citep{qiu2025agentdistill} produces MCP modules for
training-free transfer.
At the system level, ADAS~\citep{hu2024adas},
AgentSquare~\citep{shang2024agentsquare}, AFlow~\citep{zhang2024aflow}, and
MaAS~\citep{zhang2025maas} automate agentic-system or workflow design.
All start from \emph{explicit task specifications}; \skilldisco mines a
\emph{corpus of execution traces}, grounding skills in recurring behavior.
\textit{(2) Skill induction from agent experience.}
Voyager~\citep{wang2023voyager} commits one JavaScript function per
successful task, leaving near-duplicate variants unconsolidated.
SkillWeaver~\citep{zheng2025skillweaver} derives APIs from website
affordances, not workload traces.
ASI~\citep{wang2025asi} shows programs beat free-form text as skill
representations but still operates per-trajectory.
\skilldisco is an offline \emph{corpus-level compiler}: semantic clustering
merges equivalent operations into verified executable skills, so library
size scales with distinct behaviors, not episode count.

\section{Conclusion}
\label{sec:conclusion}

This paper studies procedural skill discovery by making shared execution
structure across successful traces explicit.
We formulate skills as reusable PFSM subgraphs under parameter binding, and introduce \skilldisco which distills
compact, high-coverage structures from traces and compiles them into executable, and verifiable skills. 
Experiments on \textit{ALFWorld} and \textit{WebArena} show
that these structurally grounded skills improve success rate and turn efficiency, supporting more reliable and transferable agent experience reuse.

\section*{Limitations}
\label{sec:limitations}

\textbf{Procedural tasks only.}
SkillDisCo compiles programmatic skills from interaction traces and is effective for
tasks with reusable procedural structure (navigation, web automation, tool use).
It offers no benefit for pure NLP tasks such as text generation or reading
comprehension, where success depends on linguistic understanding rather than
executable procedures.

\textbf{Pipeline quality depends on model capability.}
SkillDisCo's pipeline relies on the compiler LLM's reasoning and code-generation ability.
Insufficient model capability yields incorrect or overly specific skills, and
output quality degrades as model capability decreases..

\textbf{Requires successful traces.}
Only successful episodes contribute distillation signal. In domains where even
frontier models succeed rarely, the corpus may be too sparse for reliable
corpus-level skill extraction.

\bibliographystyle{acl_natbib}

\begin{thebibliography}{23}
\providecommand{\natexlab}[1]{#1}

\bibitem[{Cai et~al.(2023)Cai, Wang, Ma, Chen, and Zhou}]{cai2023toolmakers}
Tianle Cai, Xuezhi Wang, Tengyu Ma, Xinyun Chen, and Denny Zhou. 2023.
\newblock \href {https://arxiv.org/abs/2305.17126} {Large language models as tool makers}.
\newblock \emph{arXiv preprint arXiv:2305.17126}.

\bibitem[{Gao et~al.(2023)Gao, Madaan, Zhou, Alon, Liu, Yang, Callan, and Neubig}]{gao2023pal}
Luyu Gao, Aman Madaan, Shuyan Zhou, Uri Alon, Pengfei Liu, Yiming Yang, Jamie Callan, and Graham Neubig. 2023.
\newblock \href {https://arxiv.org/abs/2211.10435} {{PAL}: Program-aided language models}.
\newblock \emph{arXiv preprint arXiv:2211.10435}.

\bibitem[{Hu et~al.(2024)Hu, Lu, and Clune}]{hu2024adas}
Shengran Hu, Cong Lu, and Jeff Clune. 2024.
\newblock \href {https://arxiv.org/abs/2408.08435} {Automated design of agentic systems ({ADAS})}.
\newblock \emph{arXiv preprint arXiv:2408.08435}.

\bibitem[{Liang et~al.(2023)Liang, Huang, Xia, Xu, Hausman, Ichter, Florence, and Zeng}]{liang2023code}
Jacky Liang, Wenlong Huang, Fei Xia, Peng Xu, Karol Hausman, Brian Ichter, Pete Florence, and Andy Zeng. 2023.
\newblock Code as policies: Language model programs for embodied control.
\newblock In \emph{IEEE International Conference on Robotics and Automation (ICRA)}, pages 9493--9500.

\bibitem[{Liu et~al.(2023)Liu, Yu, Zhang, Xu, Lei, Lai, Gu, Ding, Men, Yang, Zhang, Deng, Zeng, Du, Zhang, Shen, Zhang, Su, Sun, Huang, Dong, and Tang}]{liu2023agentbench}
Xiao Liu, Hao Yu, Hanchen Zhang, Yifan Xu, Xuanyu Lei, Hanyu Lai, Yu~Gu, Hangliang Ding, Kaiwen Men, Kejuan Yang, Shudan Zhang, Xiang Deng, Aohan Zeng, Zhengxiao Du, Chenhui Zhang, Sheng Shen, Tianjun Zhang, Yu~Su, Huan Sun, and 3 others. 2023.
\newblock \href {https://arxiv.org/abs/2308.03688} {{AgentBench}: Evaluating {LLM}s as agents}.
\newblock \emph{arXiv preprint arXiv:2308.03688}.

\bibitem[{Ni et~al.(2026)Ni, Liu, Liu, Sun, Zhou, Cheng, Wang, Zhao, Jiang, and Jiang}]{trace2skill2026}
Jingwei Ni, Yihao Liu, Xinpeng Liu, Yutao Sun, Mengyu Zhou, Pengyu Cheng, Dexin Wang, Erchao Zhao, Xiaoxi Jiang, and Guanjun Jiang. 2026.
\newblock \href {https://arxiv.org/abs/2603.25158} {{Trace2Skill}: Distill trajectory-local lessons into transferable agent skills}.
\newblock \emph{arXiv preprint arXiv:2603.25158}.
\newblock Work in progress.

\bibitem[{Qian et~al.(2023)Qian, Han, Fung, Qin, Liu, and Ji}]{qian2023creator}
Cheng Qian, Chi Han, Yi~R. Fung, Yujia Qin, Zhiyuan Liu, and Heng Ji. 2023.
\newblock \href {https://arxiv.org/abs/2305.14318} {{CREATOR}: Tool creation for disentangling abstract and concrete reasoning of large language models}.
\newblock \emph{arXiv preprint arXiv:2305.14318}.

\bibitem[{Qiu et~al.(2025)Qiu, Juan, Wang, Yang, Qi, Zhang, Guo, Lu, Yao, Wang et~al.}]{qiu2025agentdistill}
Jieyi Qiu, Xiang Juan, Yan Wang, Lei Yang, Xin Qi, Tao Zhang, Jun Guo, Yun Lu, Zheng Yao, Mei Wang, and 1 others. 2025.
\newblock \href {https://arxiv.org/abs/2506.14728} {{AgentDistill}: Training-free agent distillation with generalizable {MCP} boxes}.
\newblock \emph{arXiv preprint arXiv:2506.14728}.

\bibitem[{Shang et~al.(2024)Shang, Li, Zhao, Ma, Liu, Xu, and Li}]{shang2024agentsquare}
Yu~Shang, Yu~Li, Keyu Zhao, Likai Ma, Jiahe Liu, Fengli Xu, and Yong Li. 2024.
\newblock \href {https://arxiv.org/abs/2410.06153} {{AgentSquare}: Automatic {LLM} agent search in modular design space}.
\newblock \emph{arXiv preprint arXiv:2410.06153}.

\bibitem[{Shinn et~al.(2023)Shinn, Cassano, Gopinath, Narasimhan, and Yao}]{shinn2023reflexion}
Noah Shinn, Federico Cassano, Ashwin Gopinath, Karthik Narasimhan, and Shunyu Yao. 2023.
\newblock Reflexion: Language agents with verbal reinforcement learning.
\newblock \emph{Advances in Neural Information Processing Systems}, 36.

\bibitem[{Shridhar et~al.(2021)Shridhar, Yuan, C\^ot\'e, Bisk, Trischler, and Hausknecht}]{shridhar2020alfworld}
Mohit Shridhar, Xingdi Yuan, Marc-Alexandre C\^ot\'e, Yonatan Bisk, Adam Trischler, and Matthew Hausknecht. 2021.
\newblock \href {https://arxiv.org/abs/2010.03768} {{ALFWorld}: Aligning text and embodied environments for interactive learning}.
\newblock In \emph{International Conference on Learning Representations}.

\bibitem[{Wang et~al.(2023)Wang, Xie, Jiang, Mandlekar, Xiao, Zhu, Fan, and Anandkumar}]{wang2023voyager}
Guanzhi Wang, Yuqi Xie, Yunfan Jiang, Ajay Mandlekar, Chaowei Xiao, Yuke Zhu, Linxi Fan, and Anima Anandkumar. 2023.
\newblock \href {https://arxiv.org/abs/2305.16291} {Voyager: An open-ended embodied agent with large language models}.
\newblock \emph{arXiv preprint arXiv:2305.16291}.

\bibitem[{Wang et~al.(2024{\natexlab{a}})Wang, Han, Ji, Wang, Baldwin, and Li}]{wang2024toolgen}
Renxi Wang, Xudong Han, Lei Ji, Shu Wang, Timothy Baldwin, and Haonan Li. 2024{\natexlab{a}}.
\newblock \href {https://arxiv.org/abs/2410.03439} {{ToolGen}: Unified tool retrieval and calling via generation}.
\newblock \emph{arXiv preprint arXiv:2410.03439}.

\bibitem[{Wang et~al.(2024{\natexlab{b}})Wang, Chen, Yuan, Zhang, Li, Peng, and Ji}]{wang2024codeact}
Xingyao Wang, Yangyi Chen, Lifan Yuan, Yizhe Zhang, Yunzhu Li, Hao Peng, and Heng Ji. 2024{\natexlab{b}}.
\newblock \href {https://arxiv.org/abs/2402.01030} {Executable code actions elicit better llm agents}.
\newblock In \emph{ICML}.

\bibitem[{Wang et~al.(2025)Wang, Gandhi, Neubig, and Fried}]{wang2025asi}
Zora~Zhiruo Wang, Apurva Gandhi, Graham Neubig, and Daniel Fried. 2025.
\newblock \href {https://arxiv.org/abs/2504.06821} {Inducing programmatic skills for agentic tasks}.
\newblock \emph{arXiv preprint arXiv:2504.06821}.

\bibitem[{Wang et~al.(2024{\natexlab{c}})Wang, Mao, Fried, and Neubig}]{wang2024awm}
Zora~Zhiruo Wang, Jiayuan Mao, Daniel Fried, and Graham Neubig. 2024{\natexlab{c}}.
\newblock \href {https://arxiv.org/abs/2409.07429} {Agent workflow memory}.
\newblock \emph{arXiv preprint arXiv:2409.07429}.

\bibitem[{Wei et~al.(2022)Wei, Wang, Schuurmans, Bosma, Ichter, Xia, Chi, Le, and Zhou}]{wei2022cot}
Jason Wei, Xuezhi Wang, Dale Schuurmans, Maarten Bosma, Brian Ichter, Fei Xia, Ed~Chi, Quoc Le, and Denny Zhou. 2022.
\newblock Chain-of-thought prompting elicits reasoning in large language models.
\newblock \emph{Advances in Neural Information Processing Systems}, 35.

\bibitem[{Yang et~al.(2024)Yang, Jimenez, Wettig, Lieret, Yao, Narasimhan, and Press}]{yang2024sweagent}
John Yang, Carlos~E Jimenez, Alexander Wettig, Kilian Lieret, Shunyu Yao, Karthik~R Narasimhan, and Ofir Press. 2024.
\newblock \href {https://arxiv.org/abs/2405.15793} {{SWE}-agent: Agent-computer interfaces enable automated software engineering}.
\newblock In \emph{The Thirty-eighth Annual Conference on Neural Information Processing Systems}.

\bibitem[{Yao et~al.(2023)Yao, Zhao, Yu, Du, Shafran, Narasimhan, and Cao}]{yao2023react}
Shunyu Yao, Jeffrey Zhao, Dian Yu, Nan Du, Izhak Shafran, Karthik Narasimhan, and Yuan Cao. 2023.
\newblock \href {https://arxiv.org/abs/2210.03629} {{ReAct}: Synergizing reasoning and acting in language models}.
\newblock In \emph{International Conference on Learning Representations}.

\bibitem[{Zhang et~al.(2025)Zhang, Niu, Fang, Wang, Bai, and Wang}]{zhang2025maas}
Guibin Zhang, Luyang Niu, Junfeng Fang, Kun Wang, Lei Bai, and Xiang Wang. 2025.
\newblock \href {https://arxiv.org/abs/2502.04180} {Multi-agent architecture search via agentic supernet}.
\newblock \emph{arXiv preprint arXiv:2502.04180}.

\bibitem[{Zhang et~al.(2024)Zhang, Xiang, Yu, Teng, Chen, Chen, Zhuge, Cheng, Hong et~al.}]{zhang2024aflow}
Jiayi Zhang, Jinyu Xiang, Zhaoyang Yu, Fengwei Teng, Xionghui Chen, Jiaqi Chen, Mingchen Zhuge, Xin Cheng, Sirui Hong, and 1 others. 2024.
\newblock \href {https://arxiv.org/abs/2410.10762} {{AFlow}: Automating agentic workflow generation}.
\newblock \emph{arXiv preprint arXiv:2410.10762}.

\bibitem[{Zheng et~al.(2025)Zheng, Fatemi, Jin, Wang, Gandhi, Song, Gu, Srinivasa et~al.}]{zheng2025skillweaver}
Boyuan Zheng, Michael~Y. Fatemi, Xiaolong Jin, Zora~Zhiruo Wang, Apurva Gandhi, Yueqi Song, Yu~Gu, Jayanth Srinivasa, and 1 others. 2025.
\newblock \href {https://arxiv.org/abs/2504.07079} {{SkillWeaver}: Web agents can self-improve by discovering and honing skills}.
\newblock \emph{arXiv preprint arXiv:2504.07079}.

\bibitem[{Zhou et~al.(2024)Zhou, Xu, Zhu, Zhou, Lo, Sridhar, Cheng, Ou, Bisk, Fried, Alon, and Neubig}]{zhou2024webarena}
Shuyan Zhou, Frank~F. Xu, Hao Zhu, Xuhui Zhou, Robert Lo, Abishek Sridhar, Xianyi Cheng, Tianyue Ou, Yonatan Bisk, Daniel Fried, Uri Alon, and Graham Neubig. 2024.
\newblock \href {https://arxiv.org/abs/2307.13854} {{WebArena}: A realistic web environment for building autonomous agents}.
\newblock In \emph{International Conference on Learning Representations}.

\end{thebibliography}

\appendix

\section{Full Results: Token Usage and Inference Cost}
\label{sec:appendix-exp-details}

\subsection{API Pricing Assumptions}
\label{sec:appendix-pricing}

All per-episode inference costs reported in the paper are computed from
provider-published token prices, retrieved from Artificial
Analysis\footnote{\url{https://artificialanalysis.ai/models\#pricing}} on
23 May 2026 (on-demand rates). Cached input tokens (returned inside a
context-caching session) are billed at $\nicefrac{1}{10}$ of the regular
input rate; output tokens are billed at the full output rate. No
batch-API discounts are applied.

\subsection{Full Baseline Comparison with Token and Cost Statistics}
\label{sec:appendix-full-baseline}

Table~\ref{tab:full-baseline-comparison} reproduces Table~\ref{tab:baseline-comparison}
from the main paper with the three token/cost columns restored.

\begin{table*}[h]
\centering
\small
\setlength{\tabcolsep}{4pt}
\caption{Full comparison against prior baselines with a GPT-4o backbone, including
per-episode token usage and inference cost. ``In-Tok.\ (Cached)'' = avg.\ input
tokens per episode with the prompt-cached portion in parentheses (cached
tokens are billed at $\nicefrac{1}{10}$ of the regular input rate).
Same convention used in all subsequent tables.}
\label{tab:full-baseline-comparison}
\begin{tabular}{@{}l c c r r r@{}}
\toprule
\textbf{Method} & \textbf{SR (\%)} $\uparrow$ & \textbf{Avg.\ Turns} $\downarrow$ & \textbf{In-Tok.\ (Cached)} & \textbf{Out-Tok.} & \textbf{Cost (\$)} \\
\midrule
\multicolumn{6}{@{}l}{\textit{(a) ALFWorld}} \\
ReAct                         & 82.00             & 19.29             & 35{,}600 (22{,}200) & 626 & 0.0450 \\
CodeAct                       & \underline{96.27} & \underline{3.63}  & \phantom{0}7{,}721 (6{,}232)   & 562 & 0.0109 \\
\cmidrule(l){2-6}
AWM\textsubscript{offline}    & 54.48             & 11.34             & 38{,}908 (22{,}923) & 275 & 0.0484 \\
ASI\textsubscript{offline}    & 47.01             & 11.43             & 68{,}945 (54{,}175) & 286 & 0.0533 \\
\cmidrule(l){2-6}
\textbf{\skilldisco\,$+$\,ReAct}
  & 92.40\,{\tiny($+12.7\%$)} & 8.78\,{\tiny($-54.5\%$)}
  & 22{,}100 (11{,}300)\,{\tiny($-37.9\%$)} & 607\,{\tiny($-3.0\%$)} & 0.0360\,{\tiny($-20.0\%$)} \\
\textbf{\skilldisco\,$+$\,CodeAct}
  & \textbf{99.25}\,{\tiny($+3.1\%$)} & \textbf{3.22}\,{\tiny($-11.3\%$)}
  & \phantom{0}8{,}468 (6{,}303)\,{\tiny($+9.7\%$)} & \textbf{368}\,{\tiny($-34.5\%$)} & \textbf{0.0107}\,{\tiny($-1.8\%$)} \\
\midrule
\multicolumn{6}{@{}l}{\textit{(b) WebArena}} \\
ReAct                         & 23.89             & 5.88              & 50{,}914 (8{,}605)  & 728 & 0.1152 \\
CodeAct                       & 19.95             & 10.86             & 75{,}962 (24{,}202) & 806 & 0.1435 \\
\cmidrule(l){2-6}
AWM\textsubscript{offline}    & 21.18             & 5.92              & \textbf{49{,}025 (11{,}748)} & 564 & \textbf{0.1018} \\
ASI\textsubscript{offline}    & \underline{24.63} & \underline{5.71}  & 56{,}662 (9{,}247)  & 603 & 0.1269 \\
\cmidrule(l){2-6}
\textbf{\skilldisco\,$+$\,ReAct}
  & \textbf{29.06}\,{\tiny($+21.6\%$)} & \textbf{5.11}\,{\tiny($-13.1\%$)}
  & 51{,}068 (8{,}582)\,{\tiny($+0.3\%$)} & \textbf{561}\,{\tiny($-22.9\%$)} & 0.1215\,{\tiny($+5.5\%$)} \\
\textbf{\skilldisco\,$+$\,CodeAct}
  & 22.91\,{\tiny($+14.8\%$)} & 8.47\,{\tiny($-22.0\%$)}
  & 85{,}649 (27{,}078)\,{\tiny($+12.8\%$)} & 736\,{\tiny($-8.7\%$)} & 0.1606\,{\tiny($+11.9\%$)} \\
\bottomrule
\end{tabular}

\vspace{2pt}
\footnotesize\raggedright
Parenthesized values give relative $\Delta$ vs.\ the matching base agent
(ReAct or CodeAct). \textbf{Best} and \underline{second-best} per column
highlighted; our rows in \textbf{bold}.
\end{table*}

\subsection{Per-Episode Token and Cost Breakdown}
\label{sec:appendix-token-detail}

Tables~\ref{tab:token-alfworld} 
expand the
aggregate ``Avg.\ In-Tok.\ (Cached) / Avg.\ Out-Tok.\ / Avg.\ Cost''
figures from Table~\ref{tab:multi-models} with the full numerical
detail for every model-setting pair.

\begin{table*}[h]
\centering
\small
\setlength{\tabcolsep}{3pt}
\caption{Per-episode token usage and inference cost on ALFWorld (CodeAct
backbone, 134 unseen tasks).
``$\Delta$'' is the relative change vs.\ the same model's Vanilla setting
(applied to In-Tok.).}
\label{tab:token-alfworld}
\begin{tabular}{@{}l r r c r r c r r c@{}}
\toprule
 & \multicolumn{3}{c}{\textbf{In-Tok.\ (Cached)}} & \multicolumn{3}{c}{\textbf{Out-Tok.}} & \multicolumn{3}{c}{\textbf{Cost (\$)}} \\
\cmidrule(lr){2-4}\cmidrule(lr){5-7}\cmidrule(lr){8-10}
\textbf{Model} & \textbf{Vanilla} & \textbf{+Skill} & \textbf{$\Delta$} & \textbf{Vanilla} & \textbf{+Skill} & \textbf{$\Delta$} & \textbf{Vanilla} & \textbf{+Skill} & \textbf{$\Delta$} \\
\midrule
Qwen3.5-4B          & 28{,}010 (24{,}566) & 19{,}608 (14{,}505) & {\scriptsize$-30.0\%$} & 2{,}095 & 1{,}391 & {\scriptsize$-33.6\%$} & 0.00049 & 0.00041 & {\scriptsize$-16.3\%$} \\
Qwen3.5-9B          & 27{,}754 (24{,}939) & 10{,}152 (\phantom{0}7{,}292) & {\scriptsize$-63.4\%$} & 1{,}382 & 610 & {\scriptsize$-55.9\%$} & 0.00061 & 0.00036 & {\scriptsize$-41.0\%$} \\
GPT-4o              & \phantom{0}7{,}721 (\phantom{0}6{,}232) & \phantom{0}8{,}468 (\phantom{0}6{,}303) & {\scriptsize$+9.7\%$}  & 562 & 368 & {\scriptsize$-34.5\%$} & 0.01090 & 0.01067 & {\scriptsize$-2.1\%$} \\
GPT-4o-mini         & 17{,}620 (16{,}020) & 13{,}686 (11{,}271) & {\scriptsize$-22.3\%$} & 891 & 485 & {\scriptsize$-45.6\%$} & 0.00102 & 0.00082 & {\scriptsize$-19.6\%$} \\
GPT-5-chat          & 10{,}606 (\phantom{0}8{,}517) & \phantom{0}7{,}728 (\phantom{0}5{,}496) & {\scriptsize$-27.1\%$} & 677 & 391 & {\scriptsize$-42.2\%$} & 0.01040 & 0.00739 & {\scriptsize$-28.9\%$} \\
GPT-5-mini          & 11{,}046 (\phantom{0}6{,}979) & \phantom{0}8{,}561 (\phantom{0}6{,}189) & {\scriptsize$-22.5\%$} & 871 & 492 & {\scriptsize$-43.5\%$} & 0.00293 & 0.00173 & {\scriptsize$-41.0\%$} \\
\bottomrule
\end{tabular}
\end{table*}

\subsection{Full Ablation Breakdown}
\label{sec:appendix-ablation-full}

Table~\ref{tab:ablation-full} expands the two-row ablation summary in
Table~\ref{tab:ablations} (\S\ref{sec:ablations}) with the no-skill
baseline and per-episode token and cost statistics.

\begin{table*}[h]
\centering
\caption{Full ablation breakdown on ALFWorld (CodeAct + GPT-4o).}
\label{tab:ablation-full}
\small\setlength{\tabcolsep}{4pt}
\begin{tabular}{@{}l r r r r r r@{}}
\toprule
\textbf{Configuration} & \textbf{\#Sk} & \textbf{SR (\%)} & \textbf{Turns} & \textbf{In-Tok.\ (Cached)} & \textbf{Out-Tok.} & \textbf{Cost (\$)} \\
\midrule
Vanilla (no skills)        & 0  & 96.27          & 3.63  & \phantom{00}7{,}720 (\phantom{0}6{,}231)  & 561 & 0.01090 \\
\midrule
A1: no distill             & 43 & 52.99          & 11.51 & 107{,}528 (97{,}901) & 994 & 0.05849 \\
A2: no compile             & 0  & 97.01          & 3.59  & \phantom{0}19{,}083 (17{,}393) & 559 & 0.01417 \\
\midrule
\textbf{Full \skilldisco}  & 5  & \textbf{99.25} & \textbf{3.22} & \phantom{00}8{,}468 (\phantom{0}6{,}302) & 368 & \textbf{0.01067} \\
\bottomrule
\end{tabular}
\end{table*}

\section{Pipeline Stage Prompts}
\label{sec:appendix-prompts}

We summarize the LLM prompts that drive each pipeline stage
(\texttt{latest\_pipeline/skill\_mining/}).  For brevity each listing
keeps the system message's role, key directives, and the JSON
output schema; long motivating examples, repeated rationale, and
formatting boilerplate from the source files are elided with
``\dots''.  Placeholders in \texttt{\{braces\}} are Python f-string
slots filled at runtime.  A few slots in Stages~4 and~5 are
specialized per benchmark (\texttt{benchmark\_prompts.py}) -- we
leave them as placeholders here and omit the benchmark-specific
fills for brevity.

\subsection{Stage 1 — Trace $\to$ Program Transpilation}
\label{sec:appendix-prompt-stage1}

\begin{lstlisting}[caption={Stage 1 prompt (trace normalization, from \texttt{prompt/trace\_converter.md}).}, label={lst:prompt-stage1}]
[System]
You are a Senior AI Systems Architect. Convert a raw ReAct trace
(alternating THOUGHT / ACTION messages and environment observations)
into a structured Python program that preserves the reasoning while
making control flow explicit and logically stable.

Each reasoning block must follow the Unified Reasoning Template:
  # A. Deterministic (logic-controlled)
  if "<condition>" in observation:
      action = "<derived action>"
      context.append((observation, None, action))
      observation, available_actions = env.step(action)
  # B. Cognitive (LLM-driven, for adaptive / uncertain steps)
  thinking, action = llm(task, context, observation, available_actions)
  context.append((observation, thinking, action))
  observation, available_actions = env.step(action)

Rules: (i) use deterministic control flow (if/for/while) for repetitive
or stabilized reasoning; (ii) reserve llm() for adaptive decisions;
(iii) update `context` and call `env.step()` after every action;
(iv) preserve every ACTION from the trace -- never drop mistakes or
recovery steps; (v) copy real observation/thinking/action strings as
inline `# ...` comments.

[User]
Please convert the following agent trace history to Python code:
{raw_history}
\end{lstlisting}

\subsection{Stage 2 — Semantic Operation Extraction}
\label{sec:appendix-prompt-stage2}

\begin{lstlisting}[caption={Stage 2 prompt.}, label={lst:prompt-stage2}]
[System]
You are an expert at analyzing agent traces represented as Python code.
Extract semantic operations: groups of consecutive steps that together
accomplish one coherent sub-goal and whose internal relationship can
be expressed as control flow (if/for/while or fixed sequence).
A good operation has clear boundaries, locality, and reusability.

For each operation emit:
  name         : snake_case verb_noun
  description  : the sub-goal
  env_steps    : every (action, observation) pair -- prefer the
                 authoritative env-step log over the code, which may
                 have dropped actions during conversion
  code_snippet : the complete code block (no truncation)
  succeeded    : true iff the final observation confirms the sub-goal
                 (e.g., "Nothing happens." => false)

Output JSON: {"task_summary": "...", "operations": [ ... ]}

[User]
## Code to Analyze
```python
{trace.code}
```
{env_log_section}
Please extract all semantic operations. The env-step log is
authoritative -- group its entries by the code's structure but
include ALL actions from the log.
\end{lstlisting}

\subsection{Stage 3 — Operation Clustering (Two Passes)}
\label{sec:appendix-prompt-stage3}

Stage~3 first runs a \emph{grouping pass} over mini-batches that
proposes candidate reusable patterns, then a \emph{consolidation pass}
that merges overlapping proposals into a minimal cluster set.

\begin{lstlisting}[caption={Stage 3a prompt -- grouping.}, label={lst:prompt-stage3a}]
[System]
You are identifying reusable operation patterns. Partition the input
operations into non-overlapping groups whose members share the largest
reusable action/observation pattern and semantic goal -- transferring
across different objects, locations, pages, or task instances.

Rules: use `env_steps` to decide membership (not names); ignore
incidental differences; split only when the action pattern or goal
differs; ignore operations that did not change state; prefer fewer,
larger, uniform groups -- when in doubt, MERGE.

Output JSON: {"groups": [{"group_name", "description",
"representative_operations": [{"operation_id","name"}]}]}

[User]
## Operations to Analyze ({len(batch)} operations)
```json
{json.dumps(batch, indent=2)}
```
\end{lstlisting}

\begin{lstlisting}[caption={Stage 3b prompt -- consolidation.}, label={lst:prompt-stage3b}]
[System]
Merge the candidate operation groups into a minimal, non-redundant
skill set. Assign every input group `index` to exactly one final
cluster (no overlap, no missing index).

Merge groups whose reusable skill, goal, or induced state change is
similar, even if the wording, objects, locations, or appliances
differ. Proceed in three steps: (1) draft candidate clusters,
(2) merge duplicates/overlaps, (3) emit the final JSON as the LAST
fenced `json` block in your response.

Output JSON: {"clusters": [{"name", "description",
"indices": [0,3,7]}, ...]}

[User]
## Groups to Consolidate ({len(groups)} groups)
```json
{json.dumps(group_summaries, indent=2)}
```
\end{lstlisting}

\subsection{Stage 4 — Skill Contract Definition}
\label{sec:appendix-prompt-stage4}

Stage~4 turns each cluster into a typed skill contract.  The base
prompt is built by \texttt{build\_stage4\_prompt(bench\_cfg)} and
contains two benchmark slots,
\texttt{\{action\_construction\_note\}} and
\texttt{\{action\_seq\_example\}}, that are filled per benchmark.

\begin{lstlisting}[caption={Stage 4 prompt (benchmark slots left as placeholders).}, label={lst:prompt-stage4}]
[System]
You design reusable skill APIs for LLM agents. Skills are Python
functions of the form
    def skill_name(param1, param2, ...):
`env` is a global -- never a parameter -- and is used as
    observation, available_actions = env.step(action: str)
After a skill returns, the caller LLM sees ONLY the returned Dict;
it must always include at least `success`, the latest `observation`
and `available_actions`, and a process trace of (action, observation).

Design principles: (1) generalize -- parameterize object names,
location lists, and action templates; (2) single responsibility --
one goal per skill; (3) action SELECTION not construction --
    {action_construction_note}
(4) declare every state change in `side_effects` and reflect it in
`canonical_action_sequence`; (5) compose freely -- assume no
prerequisite from the caller.

Output JSON contract:
  {"skill_name", "description", "docstring",
   "parameters": [{"name","type","description","required","default"}],
   "return_type": "Dict",
   "preconditions", "postconditions", "side_effects",
   "canonical_action_sequence": [
       "Ordered action templates with {param} placeholders.",
       {action_seq_example}
       "Derive ONLY from SUCCEEDED traces; prefer the shortest
        successful sequence; copy the exact action syntax."
   ],
   "abstraction_level": "primitive|composite|workflow",
   "estimated_actions_saved": <int>, "confidence_score": <float>}

[User]
## Operation Group
- Name / Description / #operations / #traces / avg_actions /
  reusability_score: {cluster fields}

## Representative Operations (real env.step() patterns)
```json
{json.dumps(examples, indent=2)}
```
Define a general, reusable skill that abstracts this pattern.
\end{lstlisting}

\subsection{Stage 5 — Skill Synthesis}
\label{sec:appendix-prompt-stage5}

Stage~5 synthesises a self-contained Python implementation from the
Stage~4 contract.  The base prompt is built by
\texttt{build\_stage5\_prompt(bench\_cfg)} and contains two benchmark
slots, \texttt{\{first\_action\_rule\}} (how to obtain the first vs.\
later actions) and \texttt{\{identifier\_note\}} (an env-specific
gotcha).

\begin{lstlisting}[caption={Stage 5 prompt (benchmark slots left as placeholders).}, label={lst:prompt-stage5}]
[System]
You synthesise the Python body of a skill defined by the Stage 4
contract. `env` is global; signature contains only domain parameters.

Critical rules:
1. Use ONLY `env.step()` and the Python stdlib -- no helper
   functions, no simulation, no fabricated observations.
2. {first_action_rule}
3. Match the trace's action syntax exactly; when constructing,
   reproduce the verb / preposition / argument pattern from the
   reference traces.
4. Branch on the LATEST `observation`, not on a self-computed flag.
5. Honour every parameter; implement every step in
   `canonical_action_sequence` and only those plus declared
   side effects.
6. Append every (action, observation) pair to `process_trace`.
7. Return a Dict including at least `success`, latest `observation`,
   latest `available_actions`, and `process_trace` (plus any extra
   keys the caller needs).

Environment note: {identifier_note}

Output JSON: {"implementation": "def skill_name(...): \\n  ...",
              "example_usage":  "result = skill_name(...) ..."}

[User]
## Skill Definition
{skill.skill_name / description / parameters / pre & postconditions}

## Representative Code Examples (ground truth env.step() patterns)
```json
{json.dumps(examples, indent=2)}
```
Synthesise the implementation. Select actions from
`available_actions`, follow the code_snippet patterns exactly, and
return the Dict so the caller can continue without blind spots.
\end{lstlisting}

\section{Skill Library Examples}
\label{sec:appendix-skill-examples}

We show one representative skill for each benchmark, reproduced verbatim
from the final compiled skill catalogs.
Listing~\ref{lst:skill-alfworld-search} is the highest-coverage skill in
the ALFWorld library: it iterates over a list of candidate locations,
opens any closed containers, and returns as soon as the target item is
observed. Listing~\ref{lst:skill-webarena-submit} is a representative
WebArena skill that drives the multi-step submission workflow on a Reddit
clone (open form $\to$ fill URL / title / body $\to$ select forum $\to$
submit), with every action selected by content-based matching against
the live admissible-action set via \texttt{find\_bid(...)}.


\begin{lstlisting}[language=Python, caption={ALFWorld skill: \texttt{systematic\_search\_locations}.}, label={lst:skill-alfworld-search}]
def systematic_search_locations(locations, target):
    """Systematically search through a list of locations to find a
    specific target item. The skill navigates to each location, opens
    it if necessary, and examines its contents to locate the target.

    Args:
        locations: A list of location names to search through.
        target: The name of the item to search for.
    """
    process_trace = []
    found_in = None

    for location in locations:
        # Step 1: Go to the location
        action = f"go to {location}"
        observation, available_actions = env.step(action)
        process_trace.append((action, observation))

        # Step 2: Check if the location is closed and open it if necessary
        if "closed" in observation:
            action = next(
                (a for a in available_actions
                 if "open" in a.lower() and location in a.lower()),
                f"open {location}",
            )
            observation, available_actions = env.step(action)
            process_trace.append((action, observation))

        # Step 3: Inspect the contents of the location
        if target in observation:
            found_in = location
            return {
                "success": True,
                "found_in": found_in,
                "observation": observation,
                "available_actions": available_actions,
                "process_trace": process_trace,
            }

    # If the target is not found in any location
    return {
        "success": False,
        "found_in": None,
        "observation": observation,
        "available_actions": available_actions,
        "process_trace": process_trace,
    }
\end{lstlisting}

\begin{lstlisting}[language=Python, caption={WebArena skill: \texttt{create\_submission\_workflow} (Reddit site).}, label={lst:skill-webarena-submit}]
def create_submission_workflow(
    title: str, 
    body: str, 
    forum_name: str,
    url: Optional[str] = None):
    
    """Complete workflow for creating a new submission/post. Navigates
    to the submission form, fills in required fields (URL if present,
    title, body), selects the target forum, and submits the form.
    Flow: click Submit -> fill fields -> select forum -> click Create.

    Args:
        title: Title of the submission to post.
        body: Main content/body of the submission.
        forum_name: Target forum/subreddit label to select.
        url: Optional URL to include if the form has a URL field.
    """
    process_trace = []
    matched_elements = []
    failure_reason = None

    observation = env.step('noop(0)')

    # Step 1: Click "Submit" to open the submission form
    submit_bid = find_bid('Submit', observation)
    if not submit_bid:
        return {'success': False, 'observation': observation,
                'process_trace': process_trace,
                'matched_elements': matched_elements,
                'failure_reason': "Submit button not found."}
    observation = env.step(f"click('{submit_bid}')")
    process_trace.append((f"click('{submit_bid}')", observation))
    matched_elements.append('Submit')

    # Step 2: Optionally fill URL field
    if url:
        url_bid = find_bid('URL', observation)
        if url_bid:
            act = f"fill({url_bid!r}, {str(url)!r})"
            observation = env.step(act)
            process_trace.append((act, observation))
            matched_elements.append('URL')

    # Step 3: Fill Title field
    title_bid = find_bid('Title', observation)
    if not title_bid:
        return {'success': False, 'observation': observation,
                'process_trace': process_trace,
                'matched_elements': matched_elements,
                'failure_reason': "Title field not found."}
    act = f"fill({title_bid!r}, {str(title)!r})"
    observation = env.step(act)
    process_trace.append((act, observation))
    matched_elements.append('Title')

    # Step 4: Open forum-selection dropdown and pick the target forum
    combobox_bid = find_bid('Choose one', observation) \
                   or find_bid('Forum', observation)
    if not combobox_bid:
        return {'success': False, 'observation': observation,
                'process_trace': process_trace,
                'matched_elements': matched_elements,
                'failure_reason': "Forum selection combobox not found."}
    observation = env.step(f"click('{combobox_bid}')")
    process_trace.append((f"click('{combobox_bid}')", observation))
    forum_bid = find_bid(forum_name, observation)
    if not forum_bid:
        return {'success': False, 'observation': observation,
                'process_trace': process_trace,
                'matched_elements': matched_elements,
                'failure_reason': f"Forum '{forum_name}' not found."}
    observation = env.step(f"click('{forum_bid}')")
    process_trace.append((f"click('{forum_bid}')", observation))
    matched_elements.append(forum_name)

    # Step 5: Fill Body field
    body_bid = find_bid('Body', observation)
    if not body_bid:
        return {'success': False, 'observation': observation,
                'process_trace': process_trace,
                'matched_elements': matched_elements,
                'failure_reason': "Body field not found."}
    act = f"fill({body_bid!r}, {str(body)!r})"
    observation = env.step(act)
    process_trace.append((act, observation))
    matched_elements.append('Body')

    # Step 6: Click "Create submission" to finalize
    create_bid = find_bid('Create submission', observation) \
                 or find_bid('Submit', observation)
    if not create_bid:
        return {'success': False, 'observation': observation,
                'process_trace': process_trace,
                'matched_elements': matched_elements,
                'failure_reason': "Create-submission button not found."}
    observation = env.step(f"click('{create_bid}')")
    process_trace.append((f"click('{create_bid}')", observation))
    matched_elements.append('Create submission')

    success = 'submitted' in observation.lower() \
              or 'created' in observation.lower()
    return {
        'success': success,
        'observation': observation,
        'process_trace': process_trace,
        'matched_elements': matched_elements,
        'failure_reason': None 
            if success
            else 'No success confirmation detected.',
    }
\end{lstlisting}

\end{document}